\documentclass[10pt,twocolumn,letterpaper]{article}

\usepackage{iccv}
\usepackage{times}
\usepackage{epsfig}
\usepackage{graphicx}
\usepackage{amsmath}
\usepackage{amssymb}

\usepackage{amsfonts}
\usepackage{multirow}
\usepackage{algorithm}
\usepackage{algorithmic}

\usepackage{subfigure}
\usepackage{amsthm}
\usepackage{url}            
\usepackage{amsfonts}       
\newcommand{\xvec}{{\bf x}}
\usepackage{graphicx}

\usepackage[switch]{lineno}
\usepackage{booktabs}

\usepackage[breaklinks=true,bookmarks=false]{hyperref}

\iccvfinalcopy 

\def\iccvPaperID{29} 

\ificcvfinal\fi

\begin{document}

\title{Towards Category and Domain Alignment: \\
Category-Invariant Feature Enhancement for Adversarial Domain Adaptation}

\author{Yuan Wu\\
Carleton University\\
{\tt\small yuan.wu3@carleton.ca}
\and
Diana Inkpen\\
University of Ottawa\\
{\tt\small  Diana.Inkpen@uottawa.ca}

\and
Ahmed El-Roby\\
Carleton University\\
{\tt\small Ahmed.ElRoby@carleton.ca}
}

\maketitle
\ificcvfinal\thispagestyle{empty}\fi

\begin{abstract}
   Adversarial domain adaptation has made impressive advances in transferring knowledge from the source domain to the target domain by aligning feature distributions of both domains. These methods focus on minimizing domain divergence and regard the adaptability, which is measured as the expected error of the ideal joint hypothesis on these two domains, as a small constant. However, these approaches still face two issues: (1) Adversarial domain alignment distorts the original feature distributions, deteriorating the adaptability; (2) Transforming feature representations to be domain-invariant needs to sacrifice domain-specific variations, resulting in weaker discriminability. In order to alleviate these issues, we propose category-invariant feature enhancement (CIFE), a general mechanism that enhances the adversarial domain adaptation through optimizing the adaptability. Specifically, the CIFE approach introduces category-invariant features to boost the discriminability of domain-invariant features with preserving the transferability. Experiments show that the CIFE could improve upon representative adversarial domain adaptation methods to yield state-of-the-art results on five benchmarks.
\end{abstract}

\section{Introduction}

In typical supervised machine learning algorithms, the training data and the test data are assumed to stem from the same distribution \cite{goodfellow2016deep,wu2021conditional}. Unfortunately, in many real-world cases, a shortage of labeled data in the interested domain is not uncommon. Thus, it is of great significance to investigate how to apply knowledge learned from a label-dense (source) domain to a label-scarce (target) domain. As the distributions of these two domains are often different, deep neural networks trained on the source domain are inclined to make spurious predictions on the target domain \cite{pan2009survey,wu2020dual,wu2021mixup}.

To address the above issue, domain adaptation is proposed to learn transferable representations across domains such that a model trained on the source domain can simultaneously perform well on the target domain \cite{wu2020dualb}. Early domain adaptation methods reweigh the source instances based on their associations to the target domain with regard to human-engineered features \cite{gong2013connecting}.  Motivated by the domain adaptation theory \cite{ben2007analysis,ben2010theory}, which suggests that the expected error on the target domain is bounded by three elements: (1) the expected error on the source domain; (2) the divergence between the two domains; (3) the adaptability. Recent methods focus on minimizing the domain divergence and explore two possible strategies for aligning different domains. The first one is to minimize some measures of domain distance, such as maximum mean discrepancy (MMD) \cite{yan2017mind} and correlation distances \cite{sun2016deep}. The second one is adversarial domain adaptation, which employs a two-player minimax game similar to generative adversarial networks (GANs) \cite{goodfellow2014generative}. In this paradigm, a domain discriminator is trained against a feature extractor, the domain discriminator aims to distinguish the source features from the target features while the feature extractor tries to confuse the discriminator. When these two components reach equilibrium, the learned features can be regarded as domain-invariant. These adversarial domain adaptation methods \cite{long2018conditional,wu2020dualb} have yielded state-of-the-art results.

For most existing domain adaptation methods, the adaptability is assumed to be a small constant that never varies in the process of domain alignment \cite{ben2010theory}. However, this assumption is often violated in practical. Given feature representations, the adaptability can be explicitly quantified as the expected error of the ideal joint hypothesis over the source and target domains. When the adaptability is poor, good domain adaptation models can not be expected. As transforming features to be domain-invariant will inevitably distort the original feature distributions and enlarge the error of the ideal joint hypothesis, a good adaptability can not be fully guaranteed. Moreover, in the process of learning domain-invariant features, the transferability is enhanced at the expense of sacrificing discriminability \cite{chen2019transferability}. In this paper, we propose a novel category-invariant feature enhancement (CIFE) mechanism, which introduces category-invariant features, to address the above two issues. Similar to domain alignment, the generation of category-invariant features can also be formulated as a two-player game: a feature extractor is trained against a category discriminator, the category discriminator tries to distinguish different labels, while the feature extractor aims to fool the category discriminator. By adversarially training the feature extractor and the category discriminator, we can make the learned features transferable across categories, i.e. category-invariant. We term this process as category alignment. The category-invariant features are supposed to represent domain-specific information and boost the model adaptability by complementing the discriminability of the domain-invariant features. Experiments show that our method enables existing adversarial domain adaptation models to learn transferable feature representations without sacrificing much discriminability, and yield state-of-the-art results on five benchmarks. The contributions of our paper are summarized as follows: 

\begin{itemize}
    \item We propose a category-invariant feature enhancement (CIFE) mechanism, which enhances the discriminability of the domain-invariant features by introducing the category-invariant features. The proposed CIFE improves the system performance by optimizing the adaptability, rather than further reducing the domain divergence.
    
    \item To evaluate the efficacy of CIFE, we embed CIFE into two existing adversarial domain adaptation methods and evaluate them on five benchmarks. Our proposed CIFE significantly improves upon these two methods by yielding state-of-the-art results. 
    
    \item Further experiments are conducted to validate the feasibility of advancing domain adaptation by optimizing the adaptability, and explore how the hyperparameter influences the performance of the model.
\end{itemize}

\section{Related Work}

The main objective of domain adaptation is to transfer the knowledge learned from the source domain to the target domain. Unsupervised domain adaptation (UDA) tackles a more challenging scenario where there is no direct access to the label information of the target domain. As deep neural networks can automatically extract feature representations from massive data, deep neural network-based methods have been widely studied for UDA. The deep adaptation network (DAN) applies maximum mean discrepancy (MMD) to layers embedded in a reproducing kernel hilbert space, effectively matching higher-order statistics of the two distributions \cite{long2015learning}. The joint adaptation network (JAN) learns a learner by aligning the joint distributions of multiple domain-specific layers across different domains based on a joint MMD criterion \cite{long2017deep}. The deep correlation alignment (CORAL) proposes to match the means and covariances of two domains \cite{sun2016deep}.  

Inspired by the success of generative adversarial networks (GANs), \cite{ganin2016domain} proposes domain discriminative neural networks (DANNs) which could learn domain-invariant features by deploying adversarial learning between a domain discriminator and a feature extractor. DANN projects the source and target domains into a shared latent space and adversarially performs domain alignment to reduce the domain divergence in the domain adaptation theory \cite{ben2010theory}. Based on adversarial domain adaptation, a line of works further reduce the domain divergence through improving the domain discriminator or the procedure of adversarial learning. The adversarial discriminative domain adaptation (ADDA) uses asymmetric feature extractors for the two domains to conduct the alignment \cite{tzeng2017adversarial}. The multi-adversarial domain adaptation (MADA) captures multi-mode structures by re-weighting features with category predictions \cite{pei2018multi}. The cycle-consistent adversarial domain adaptation (CyCADA) implements domain adaptation at both pixel-level and feature-level by using cycle-consistent adversarial training \cite{hoffman2018cycada}. The conditional adversarial domain adaptation (CDAN) conditions the domain discriminator on discriminative information by multiplicative interactions between feature representations and predictions \cite{long2018conditional}. The batch spectral penalization (BSP) penalizes the largest singular values to strengthen other eigenvectors of the learned domain-invariant features to boost the discriminability \cite{chen2019transferability}. The dynamic adversarial adaptation network (DAAN) dynamically learns domain-invariant representations while quantitatively evaluating the relative importance of global and local domain distributions \cite{yu2019transfer}. The batch nuclear-norm maximization (BNM) enlarges the nuclear-norm of the batch output matrix to make the learned domain-invariant features more discriminative \cite{cui2020towards}. The enhanced transport distance (ETD) exploits the attention scores estimating the similarity between samples to weigh the transport distance between domains and learned discriminative features by reducing the weighed distance \cite{li2020enhanced}. The label propagation with augmented anchors (A$^2$LP) improves the label propagation via generation of unlabeled virtual samples with high confidence label prediction \cite{zhang2020label}

Most previous UDA methods concentrate on minimizing the domain divergence, few works explore optimizing the adaptability to improve the generalization ability of the model. In this paper, our CIFE approach tries to improve domain adaptation methods through optimizing the adaptability.

\section{Method}

In this work, we consider unsupervised domain adaptation in the following setting. There exist abundant labeled instances in the source domain, $D^s=\{(\xvec_i^s,y_i^s)\}_{i=1}^{n_s}$ with $\xvec_i^s\in\mathcal{X}$ and $y_i^s\in\mathcal{Y}$, and a set of unlabeled instances in the target domain, $D^t=\{\xvec_i^t\}_{i=1}^{n_t}$ with $\xvec_i^t\in\mathcal{X}$. The data in the two domains are drawn from different distributions $\mathcal{S}$ and $\mathcal{T}$, but share the same label space. The main objective is to learn a model $h:\mathcal{X}\to\mathcal{Y}$ that has a good capacity of generalizing on both source and target domains.

\subsection{Adversarial Domain Adaptation}

The key idea of adversarial domain adaptation is to learn domain-invariant features that can be generalized across domains. Starting from the domain adversarial neural networks (DANNs) \cite{ganin2016domain}, adversarial learning has been widely adopted to learn feature representations to bridge the domain divergence in UDA. Considering DANN as an example, the base network is composed of three components: a feature extractor $F$, a task-specific classifier $C$, and a binary domain discriminator $D$. The feature extractor $F:\mathcal{X}\to\mathbb{R}^m$ maps one input instance $\xvec$ from the input space $\mathcal{X}$ into a shared latent space $F(\xvec)\in\mathbb{R}^m$. The classifier $C:\mathbb{R}^m\to\mathcal{Y}$ transforms a feature vector in the shared latent space to the label space $\mathcal{Y}$. The domain discriminator $D:\mathbb{R}^m\to[0,1]$ separates the source features (with domain index 0) from the target ones (with domain index 1) in the latent space. By adversarially training $F$ to confuse $D$, DANN can learn transferable features across domains. Moreover, the feature extractor $F$ and the classifier $C$ are trained simultaneously to minimize the classification error on the source labeled data. This makes the learned features discriminative across categories. Formally, DANN can be formulated as:

\begin{align}
    \min_{F,C}\max_{D}\mathcal{L}_c(F,C)+\lambda_d\mathcal{L}_{d}(F,D)
\end{align}

\begin{align}
    \mathcal{L}_c(F,C)=\mathbb{E}_{(\xvec^s,y^s)\sim\mathcal{S}} \ell(C(F(\xvec^s)),y^s)
\end{align}

\begin{equation}
    \begin{aligned}
    \mathcal{L}_{d}(F,D)=&\mathbb{E}_{\xvec^s\sim \mathcal{S}}\log[D(F(\xvec^s))] \\
    +&\mathbb{E}_{\xvec^t\sim\mathcal{T}}\log[1-D(F(\xvec^t))]
    \end{aligned}
\end{equation}

\noindent Where $\ell(\cdot,\cdot)$ is the canonical cross-entropy loss function, and $\lambda_d$ is a trade-off hyperparameter.

\subsection{Category-Invariant Feature}

\begin{figure}
    \centering
    \includegraphics[width=0.9\columnwidth]{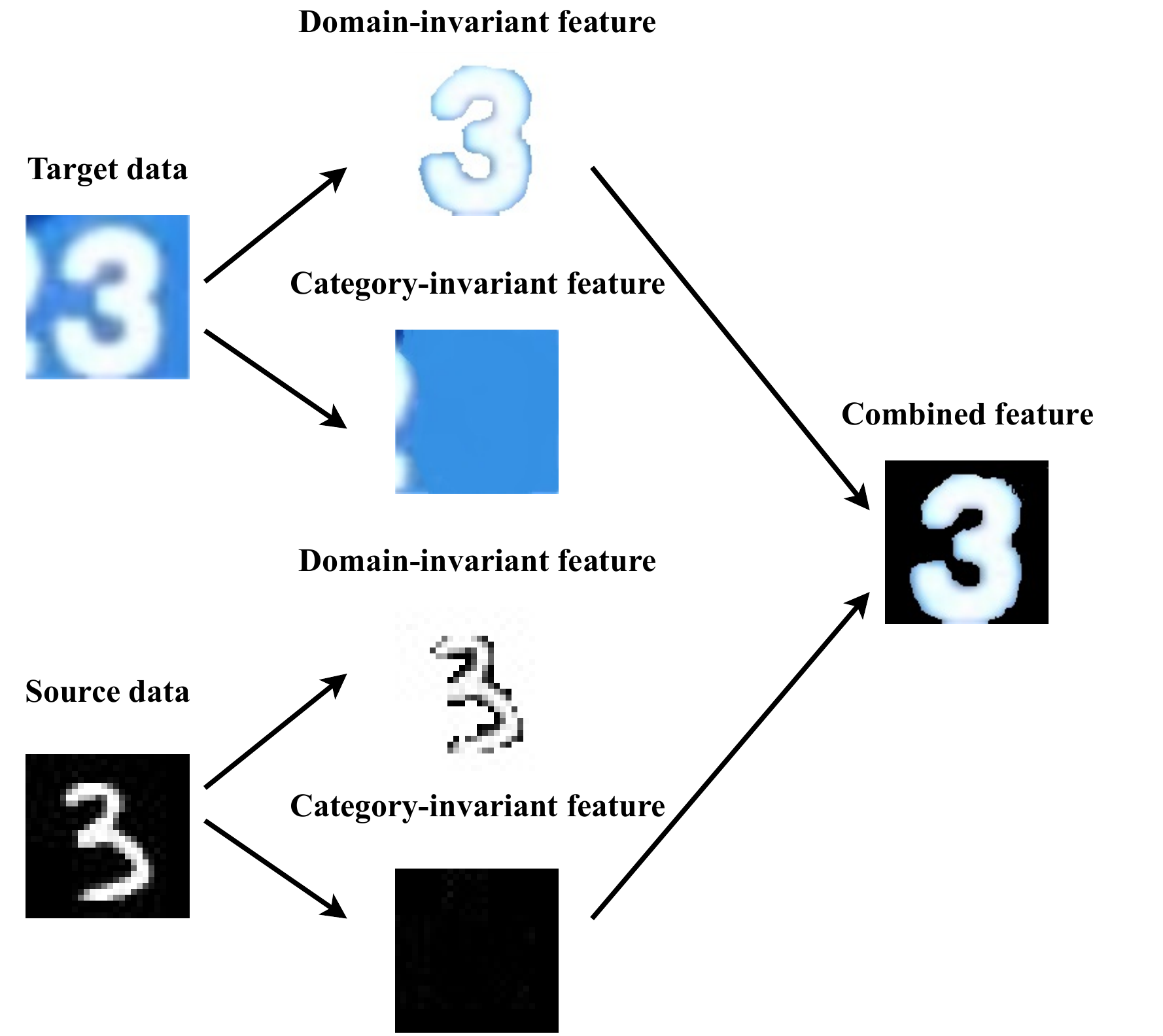}
    \caption{The illustration of the CIFE mechanism.
	}
    \label{Fig4}
\end{figure}

\begin{figure*}
    \centering
    \includegraphics[width=1.0\columnwidth]{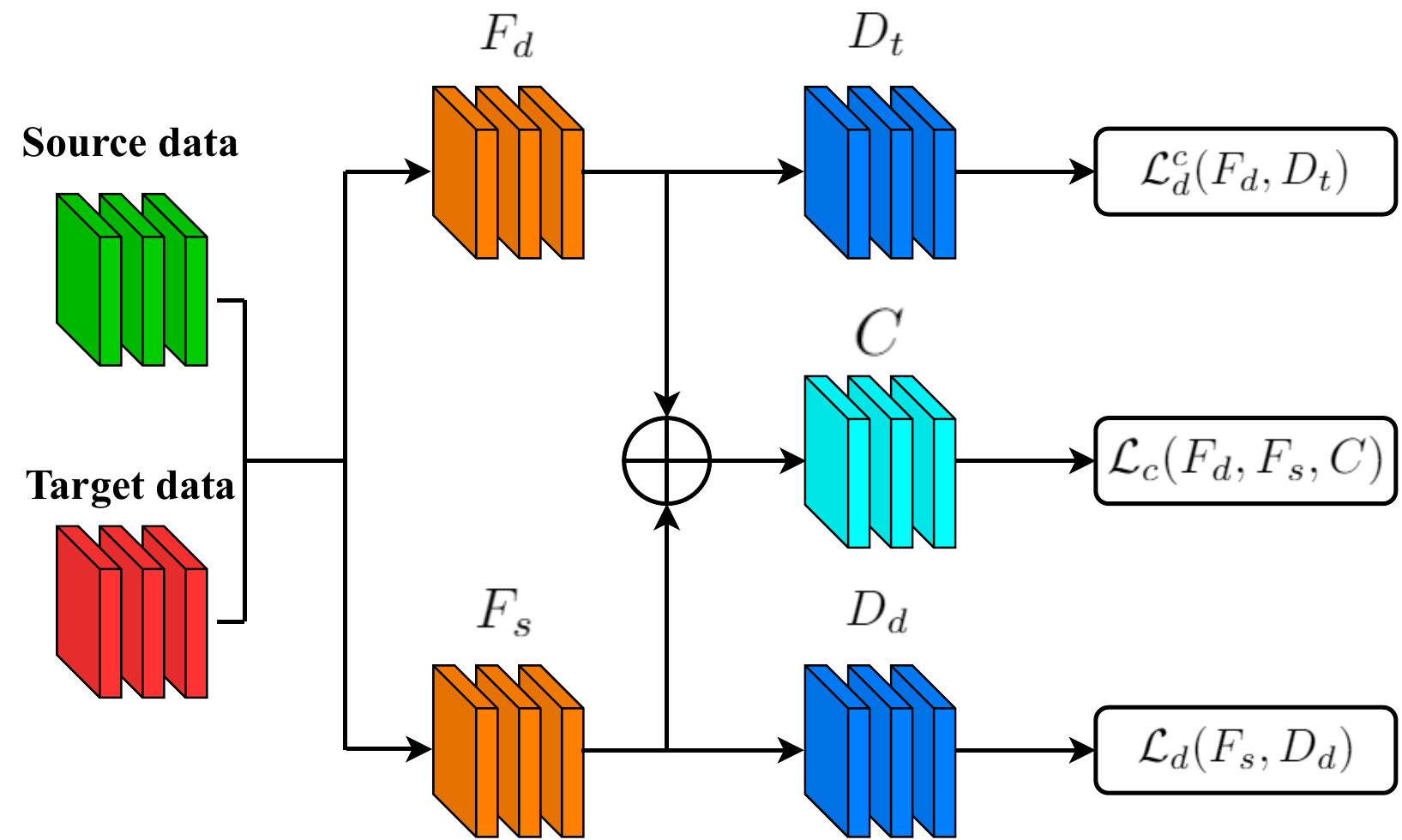}
    \caption{The architecture of CIFE+DANN where CIFE introduces category-invariant features to enhance the discriminability of the domain-invariant features learned by DANN. CIFE can be easily plugged into any adversarial domain adaptation method, which is end-to-end trainable. The CIFE+DANN model consists of five components: the domain-specific feature extractor $F_d$, which is used to capture the category-invariant features; the domain-invariant feature extractor $F_s$, which aims to learn the domain-invariant features; the category discriminator $D_t$, which identifies different labels of the input features; the domain discriminator $D_d$, which differentiates source features from target features; the classifier $C$, which is used to conduct the classification. $\mathcal{L}_c(F_d,F_s,C)$ is the typical cross-entropy loss, $\mathcal{L}_d(F_s,D_d)$ and $\mathcal{L}_d^c(F_d,D_t)$ are adversarial loss functions that guide the domain-invariant feature generation and the category-invariant feature generation, respectively.
	}
    \label{Fig1}
\end{figure*}


The batch spectral penalization \cite{chen2019transferability} reveals that the feature representations can be decomposed into eigenvectors with importance quantified by the corresponding singular values from a spectral analysis viewpoint. The feature transferability mainly resides in the eigenvectors with top singular values, the eigenvectors with low singular values embody domain-specific variations and should be discouraged. In contrast, the feature discriminability depends on all eigenvectors because the rich discriminative structures can not be fully expressed by only a few eigenvectors. Therefore, there exists a contraction between transferability and discriminability. In the process of learning transferable feature representations, we need to sacrifice some discriminability by suppressing domain-specific variations. As the target domain has no label, this sacrifice mainly resides in the target data. In order to complement the sacrificed domain-specific information, we introduce the category-invariant features. The category-invariant features can be yielded through category alignment and are supposed to be category-irrelevant and domain-specific. In practice, by replacing the domain discriminator $D$ with a category discriminator $D_t$, we can use adversarial training between the feature extractor $F$ and the category discriminator $D_t$ to obtain the category-invariant features by processing the training instances and their corresponding labels. In UDA, as we have no access to the label information of the target domain, the category-invariant features can only be learned from the source domain, which can be encoded as follows:

\begin{equation}
    \begin{aligned}
    \mathcal{L}_{d}^c(F,D_t)=\mathbb{E}_{(\xvec^s,y^s)\sim \mathcal{S}}\ell(D_t(F(\xvec^s)),y^s)
    \end{aligned}
\end{equation}

\noindent  As the category-invariant features are domain-specific, which has been demonstrated to be beneficial to their own domain and detrimental to other domains \cite{chen2018multinomial}, it is challenging to apply the source category-invariant features to boost the classification accuracy of the target data.

\subsection{Category-Invariant Feature Enhancement}

In this paper, we assume that the essential information of an image can be decomposed into two independent sets: (1) the domain-invariant information, that is discriminative across domains; (2) the domain-specific information, that is transferable across categories for one domain. For one specific image, the combination of these two types of information is expected to represent each individual characteristic of that image. For a source image and a target image that share the same label, we can obtain their domain-invariant features and category-invariant features. If these features contain no noise, e.g. the domain-invariant features are not contaminated by the domain-specific information while the category-invariant features carry no discriminative information. When we combine the category-invariant features of the source image and the domain-invariant features of the target image, the combination is expected to contain all essential information of the source image, as illustrated in Figure \ref{Fig4}. Therefore, it is feasible to feed the combinations of source category-invariant features and target domain-invariant features to a classifier trained on the source domain and obtain accurate predictions.

As stated above, the source category-invariant features can be used to complement the target domain-invariant features. Our proposed CIFE can easily be embedded into existing adversarial domain adaptation approaches, such as DANN \cite{ganin2016domain} and conditional adversarial domain adaptation (CDAN) \cite{long2018conditional}. The architecture of CIFE+DANN is illustrated in Figure \ref{Fig1}, which consists of five components: a domain-specific feature extractor $F_d$, a category discriminator $D_t$, a domain-invariant feature extractor $F_s$, a domain discriminator $D_d$, and a classifier $C$. The domain-specific feature extractor $F_d$ aims to learn the category-invariant features, while the domain-invariant feature extractor $F_s$ captures the domain-invariant features. With the CIFE applied to the DANN, there exist two two-player minimax games: The first one is played between the $F_d$ and the $D_t$, aiming to extract category-invariant features from the source domain; The second one is the typical adversarial learning, which is deployed between the $F_s$ and the $D_d$, trying to capture domain-invariant features from both the source and target domains. By incorporating these two types of adversarial learning, the transferability of the domain-invariant features can be preserved as much as possible, which is originally tailored for the domain adaptation. While the discriminability of the domain-invariant features can be enhanced by complementing with domain-specific information. The source data should be involved in both minimax objectives, while the target data only participate in the domain alignment. In our work, we concatenate category-invariant features and domain-invariant features, feeding the concatenations to the classifier $C$ as the input to conduct classification. Thus, the classification loss should be rewritten as:

\begin{equation}
    \begin{aligned}
    \mathcal{L}_c(F_d,F_s,C)=\mathbb{E}_{(\xvec^s,y^s)\sim \mathcal{S}}\ell(C([F_d(\xvec^s),F_s(\xvec^s)]),y^s)
    \end{aligned}
\end{equation}

\noindent where $[\cdot,\cdot]$ indicates the concatenation of two vectors. To learn representations with both transferability and discriminability, the dual adversarial learning of CIFE+DANN is formulated as:

\begin{equation}
    \begin{aligned}
    \min_{F_d,F_s,C}\max_{D_t,D_d}&\mathcal{L}_c(F_d,F_s,C)+\lambda_d\mathcal{L}_d(F_s,D_d) +\lambda_c\mathcal{L}_d^c(F_d,D_t)
    \end{aligned}
\end{equation}

\noindent where $\lambda_d$ and $\lambda_c$ are hyperparameters that trade-off different loss functions.

\begin{figure*}
    \centering
    \includegraphics[width=1.0\columnwidth]{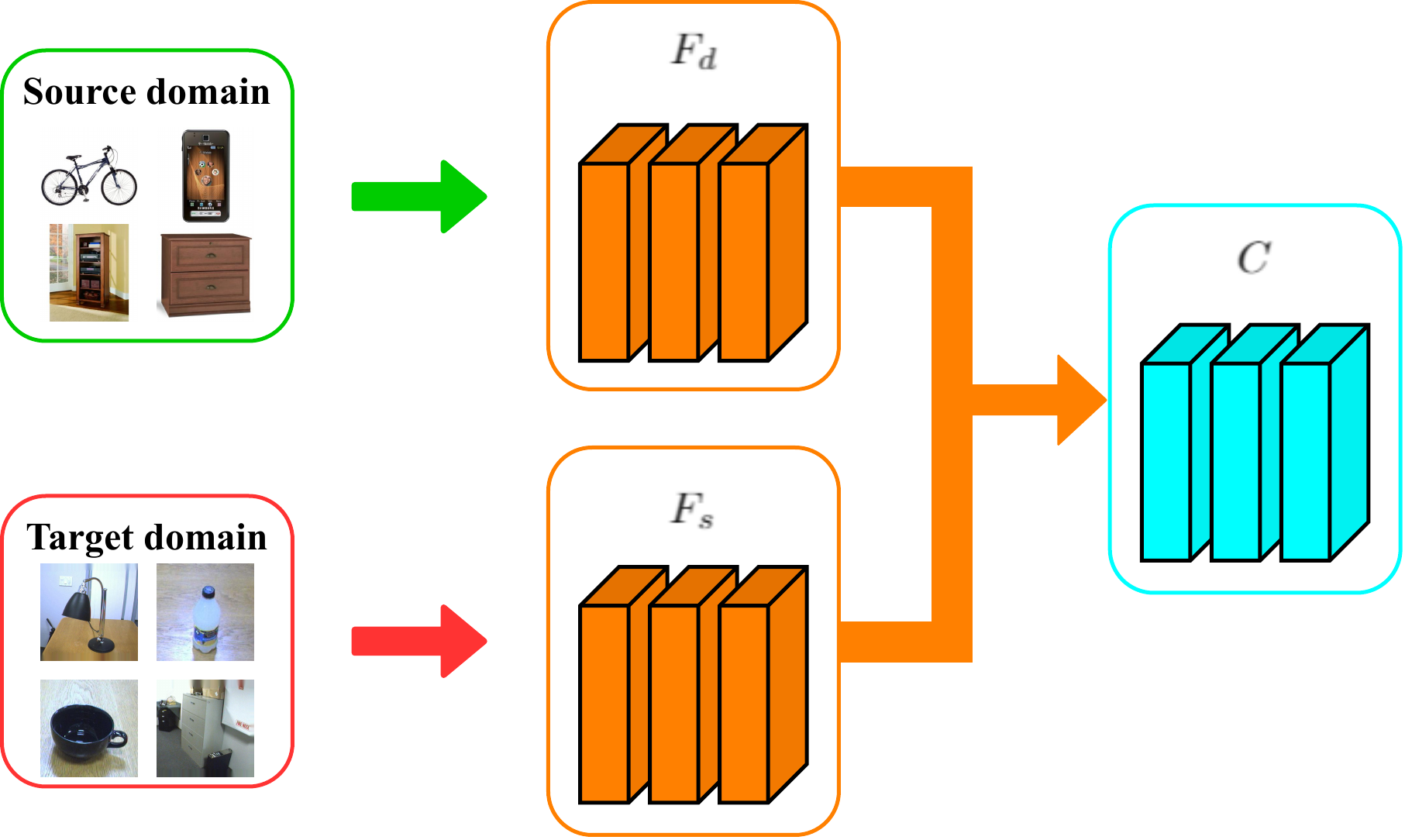}
    \caption{The predicting process of the CIFE+DANN method.
	}
    \label{Fig2}
\end{figure*}


\begin{algorithm}
\caption{Stochastic gradient descent training algorithm of CIFE+DANN}\label{trainingalg}
\begin{algorithmic}[1]
\STATE{\bf Input:} Source domain: $D^s$, target domain: $D^t$, and batch size: N.
\STATE{\bf Output:} Configurations of CIFE+DANN
\STATE{\bf Initialize} $\lambda_d$ and $\lambda_c$
\FOR{number of training iterations}
    \STATE $(\xvec^s, y^s)$ $\leftarrow$ RANDOMSAMPLE($D^s, N$)
    \STATE $(\xvec^t)$ $\leftarrow$ RANDOMSAMPLE($D^t, N$)
	\STATE Calculate $l_D$ =$\lambda_d\mathcal{L}_{d}(F_s,D_d)$+$\lambda_c\mathcal{L}_{d}^c(F_d,D_t)$;\\
	Update $D_d$ and $D_t$ by ascending along gradients $\nabla l_D$.\\[.2ex] 
	\STATE Calculate $loss$=$\mathcal{L}_c(F_d, F_s,C)$+$\lambda_d\mathcal{L}_d(F_s, D_d)$ \\
	\quad \quad \quad \quad \quad \quad+$\lambda_c\mathcal{L}_d^c(F_d,D_t)$;\\
	Update $F_s$, $F_d$ and $C$ by descending along gradients $\nabla loss$.\\[.2ex]
\ENDFOR
\end{algorithmic}
\end{algorithm}


\subsection{Training and Predicting Procedure}

The training algorithm of CIFE+DANN, which uses mini-batch stochastic gradient descent, is presented in Algorithm \ref{trainingalg}. In each iteration, the source and target samples are fed into the model to generate category-invariant and domain-invariant features. The category-invariant features are obtained by adversarially training the domain-specific feature extractor $F_d$ and the category discriminator $D_t$, while the domain-invariant features are yielded by letting the domain-invariant feature extractor $F_s$ compete with the domain discriminator $D_d$. The concatenations of category-invariant features and domain-invariant features are supposed to be both transferable and discriminative. $\lambda_d$ and $\lambda_c$ are hyperparameters balancing different losses. When evaluating the target test data, some source data are required to provide category-invariant features. Therefore, we randomly draw samples from the source training set to fulfill this requirement. The predicting process is illustrated in Figure \ref{Fig2}. 

\subsection{Discussion}

For the representation learning of adversarial domain adaptation, both transferability and discriminability are crucial. Specifically, transferability can guarantee the knowledge learned from the source domain to generalize to the target domain and discriminability enables the model to identify different categories \cite{pan2009survey,chen2019transferability}. Based on the domain adaptation theory \cite{ben2010theory}, an essential prerequisite for domain adaptation is a good adaptability over the source and target domains. In domain adaptation, there exist three main research issues: (1) what to transfer, (2) how to transfer, and (3) when to transfer \cite{pan2009survey}. Adversarial domain adaptation approaches focus on the first challenge, trying to learn both transferable and discriminative features through minimizing domain divergence adversarially. However, as stated in \cite{chen2019transferability}, this class of techniques is risky in transforming features to be domain-invariant as it needs to suppress domain-specific variations of the original feature distributions, imposing detrimental effects to the discriminability of the learned features. In addition, most previous UDA methods regard the adaptability as a small constant and ignore its influence in domain alignment, few works investigate optimizing adaptability to improve the system performance.

In this paper, we introduce category-invariant features, that are obtained by performing category alignment. These features are supposed to be category-irrelevant and domain-specific, and can be applied to compensate for the sacrificed domain-specific variations of the domain-invariant features, boosting their discriminability. By incorporating category-invariant features with domain-invariant features, the adaptability can be essentially controlled to be small in the training process, yielding lower expected error on the target domain. In summary, our proposed CIFE can enable existing adversarial domain adaptation approaches to learn both transferable and discriminative feature representations, and the category-invariant features can make contributions to the "what to transfer" research issue.

\section{Experiments}
\setlength{\tabcolsep}{4pt}
\begin{table*}
\begin{center}
\caption{Accuracy (\%) on Office-31.}
\label{table1}
    \centering
    \resizebox{0.7\linewidth}{!}{
    \smallskip \begin{tabular}{ c c c c c c c c c}
        \toprule[1pt]
        Method & A$\rightarrow$W & D$\rightarrow$W & W$\rightarrow$D & A$\rightarrow$D & D$\rightarrow$A & W$\rightarrow$A & Avg \\
        \hline
        ResNet-50 \cite{he2016deep} & 68.4$\pm$0.2 & 96.7$\pm$0.1 & 99.3$\pm$0.1 & 68.9$\pm$0.2 & 62.5$\pm$0.3 & 60.7$\pm$0.3 & 76.1 \\  
        DAN \cite{long2015learning} & 80.5$\pm$0.4 & 97.1$\pm$0.2 & 99.6$\pm$0.1 & 78.6$\pm$0.2 & 63.6$\pm$0.3 & 62.8$\pm$0.2 & 80.4 \\
        DANN \cite{ganin2015unsupervised} & 82.0$\pm$0.4 & 96.9$\pm$0.2 & 99.1$\pm$0.1 & 79.7$\pm$0.4 & 68.2$\pm$0.4 & 67.4$\pm$0.5 & 82.2 \\
        JAN \cite{long2017deep} & 85.4$\pm$0.3 & 97.4$\pm$0.2 & 99.8$\pm$0.2 & 84.7$\pm$0.3 & 68.6$\pm$0.3 & 70.0$\pm$0.4 & 84.3 \\
        MADA \cite{pei2018multi} & 90.0$\pm$0.1 & 97.4$\pm$0.1 & 99.6$\pm$0.1 & 87.8$\pm$0.2 & 70.3$\pm$0.3 & 66.4$\pm$0.3 & 85.2 \\
        CDAN \cite{long2018conditional} & 93.1$\pm$0.2 & 98.2$\pm$0.2 & \textbf{100.0}$\pm$0.0 & 89.8$\pm$0.3 & 70.1$\pm$0.4 & 68.0$\pm$0.4 & 86.6 \\
        BSP \cite{chen2019transferability} & 93.3$\pm$0.2 & 98.2$\pm$0.2 & \textbf{100.0}$\pm$0.0 & 93.0$\pm$0.2 & 73.6$\pm$0.3 & 72.6$\pm$0.3 & 88.5 \\
        \hline
        ETD \cite{li2020enhanced} & 92.1 & \textbf{100.0} & \textbf{100.0} & 88.0 & 71.0 & 67.8 & 86.2 \\
        A$^2$LP \cite{wu2020dual} & 87.7 & 98.1 & 99.0 & 87.8 & 75.8 & \textbf{75.9} & 87.4 \\
        BNM \cite{cui2020towards} & 92.8 & 98.8 & \textbf{100.0} & 92.9 & 73.5 & 73.8 & 88.6 \\
        \hline
        \textbf{CIFE+DANN} & 90.7$\pm$0.3 & 99.0$\pm$0.1 & \textbf{100.0}$\pm$0.0 & 90.0$\pm$0.5 & 71.0$\pm$0.3 & 69.9$\pm$0.3 & 86.8 \\
        \textbf{CIFE+CDAN} & \textbf{94.0}$\pm$0.2 & 99.3$\pm$0.1 & \textbf{100.0}$\pm$0.0 & \textbf{93.4}$\pm$0.2 & \textbf{75.9}$\pm$0.2 & 74.3$\pm$0.3 & \textbf{89.5} \\
        \bottomrule[1pt]
    \end{tabular}}
\end{center}
\end{table*}



\begin{table*}
\caption{Accuracy (\%) on ImageCLEF-DA.}
\label{table2}
    \centering
    \resizebox{0.7\linewidth}{!}{
    \smallskip \begin{tabular}{ c c c c c c c c c}
        \toprule[1pt]
        Method & I$\rightarrow$P & P$\rightarrow$I & I$\rightarrow$C & C$\rightarrow$I & C$\rightarrow$P & P$\rightarrow$C & Avg \\
        \hline
        ResNet-50 \cite{he2016deep} & 74.8$\pm$0.3 & 83.9$\pm$0.1 & 91.5$\pm$0.3 & 78.0$\pm$0.2 & 65.5$\pm$0.3 & 91.2$\pm$0.3 & 80.7 \\  
        DAN \cite{long2015learning} & 74.5$\pm$0.4 & 82.2$\pm$0.2 & 92.8$\pm$0.2 & 86.3$\pm$0.4 & 69.2$\pm$0.4 & 89.8$\pm$0.4 & 82.5 \\
        DANN \cite{ganin2015unsupervised} & 75.0$\pm$0.6 & 86.0$\pm$0.3 & 96.2$\pm$0.4 & 87.0$\pm$0.5 & 74.3$\pm$0.5 & 91.5$\pm$0.6 & 85.0 \\
        JAN \cite{long2017deep} & 76.8$\pm$0.4 & 88.0$\pm$0.2 & 94.7$\pm$0.2 & 89.5$\pm$0.3 & 74.2$\pm$0.3 & 91.7$\pm$0.3 & 85.8 \\
        MADA \cite{pei2018multi} & 75.0$\pm$0.3 & 87.9$\pm$0.2 & 96.0$\pm$0.3 & 88.8$\pm$0.3 & 75.2$\pm$0.2 & 92.2$\pm$0.3 & 85.8 \\
        CDAN \cite{long2018conditional} & 76.7$\pm$0.3 & 90.6$\pm$0.3 & 97.0$\pm$0.4 & 90.5$\pm$0.4 & 74.5$\pm$0.3 & 93.5$\pm$0.4 & 87.1 \\
        \hline
        DAAN \cite{yu2019transfer} & 78.5 & 91.3 & 94.4 & 88.4 & 74.0 & 94.3 & 86.8 \\
        ETD \cite{li2020enhanced} & \textbf{81.0} & 91.7 & 97.9 & 93.3 & \textbf{79.5} & 95.0 & 89.7 \\
        A$^2$LP \cite{zhang2020label} & 79.3 & 91.8 & 96.3 & 91.7 & 78.1 & 96.0 & 88.9 \\
        \hline
        \textbf{CIFE+DANN} & 77.0$\pm$0.2 & 91.1$\pm$0.2 & 97.3$\pm$0.3 & 90.8$\pm$0.3 & 74.5$\pm$0.5 & 93.7$\pm$0.3 & 87.4 \\
        \textbf{CIFE+CDAN} & 79.5$\pm$0.3 & \textbf{93.0}$\pm$0.2 & \textbf{98.2}$\pm$0.3 & \textbf{93.6}$\pm$0.3 & 79.2$\pm$0.4 & \textbf{96.1}$\pm$0.4 & \textbf{90.0} \\
        \bottomrule[1pt]
    \end{tabular}}
\end{table*}

\begin{table*}
\caption{Accuracy (\%) on Digits and VisDA-2017.}
\label{table3}
    \centering
    \resizebox{0.7\linewidth}{!}{
    \smallskip \begin{tabular}{ c c c c c | c c }
    \toprule[1pt]
    Method & M$\rightarrow$U & U$\rightarrow$M & S$\rightarrow$M & Avg & Method & Synthetic$\rightarrow$Real\\
    \hline
    No Adaptation \cite{hoffman2018cycada} & 82.2 & 69.6 & 67.1 & 73.0 & ResNet-101 \cite{he2016deep} & 52.4 \\
    DANN \cite{ganin2015unsupervised} & 90.4 & 94.7 & 84.2 & 89.8 & DANN \cite{ganin2015unsupervised} & 57.4 \\ 
    ADDA \cite{tzeng2017adversarial} & 89.4 & 90.1 & 86.3 & 88.6 & DAN \cite{long2015learning} & 61.1 \\
    CyCADA \cite{hoffman2018cycada} & 95.6 & 96.5 & 90.4 & 94.2 & JAN \cite{long2017deep} & 65.7\\
    CDAN \cite{long2018conditional} & 93.9 & 96.9 & 88.5 & 93.1 & CDAN \cite{long2018conditional} & 73.7 \\
    BSP \cite{chen2019transferability} & 95.0 & 98.1 & 92.1 & 95.1 & BSP \cite{chen2019transferability} & 75.9 \\
    \hline
    \textbf{CIFE+DANN} & 93.7 & 97.4 & 93.1 & 94.7 & \textbf{CIFE+DANN} & 74.4\\
    \textbf{CIFE+CDAN} & \textbf{96.1} & \textbf{98.8} & \textbf{94.3} & \textbf{96.4} & \textbf{CIFE+CDAN} & \textbf{78.1}\\
    \bottomrule[1pt]     
    \end{tabular}}
\end{table*}


\begin{table*}
\caption{Accuracy (\%) on Office-Home.}
\label{table4}
    \centering
    \resizebox{1.0\linewidth}{!}{
    \smallskip \begin{tabular}{c c c c c c c c c c c c c c}
    \toprule[1pt]
    Method & Ar$\rightarrow$Cl & Ar$\rightarrow$Pr & Ar$\rightarrow$Rw & Cl$\rightarrow$Ar & Cl$\rightarrow$Pr & Cl$\rightarrow$Rw & Pr$\rightarrow$Ar & Pr$\rightarrow$Cl & Pr$\rightarrow$Rw & Rw$\rightarrow$Ar & Rw$\rightarrow$Cl & Rw$\rightarrow$Pr & Avg \\
        \hline
        ResNet-50 \cite{he2016deep} & 34.9 & 50.0 & 58.0 & 34.7 & 41.9 & 46.2 & 38.5 & 31.2 & 60.4 & 53.9 & 41.2 & 59.9 & 46.1 \\  
        DAN \cite{long2015learning} & 43.6 & 57.0 & 67.9 & 45.8 & 56.5 & 60.4 & 44.0 & 43.6 & 67.7 & 63.1 & 51.5 & 74.3 & 56.3 \\
        DANN \cite{ganin2015unsupervised} & 45.6 & 59.3 & 70.1 & 47.0 & 58.5 & 60.9 & 46.1 & 43.7 & 68.5 & 63.2 & 51.8 & 76.8 & 57.6 \\
        JAN \cite{long2017deep} & 45.9 & 61.2 & 68.9 & 50.4 & 59.7 & 61.0 & 45.8 & 43.4 & 70.3 & 63.9 & 52.4 & 76.8 & 58.3 \\
        CDAN \cite{long2018conditional} & 49.0 & 69.3 & 74.5 & 54.4 & 66.0 & 68.4 & 55.6 & 48.3 & 75.9 & 68.4 & 55.4 & 80.5 & 63.8 \\
        BSP \cite{chen2019transferability} & 52.0 & 68.6 & 76.1 & 58.0 & 70.3 & 70.2 & \textbf{58.6} & 50.2 & 77.6 & \textbf{72.2} & \textbf{59.3} & 81.9 & 66.3 \\
        \hline
        \textbf{CIFE+DANN} & 47.8 & 65.9 & 73.4 & 48.3 & 62.7 & 64.2 & 48.7 & 46.9 & 74.5 & 68.2 & 53.4 & 80.7 & 61.2 \\
        \textbf{CIFE+CDAN} & \textbf{52.3} & \textbf{71.0} & \textbf{78.3} & \textbf{58.9} & \textbf{71.8} & \textbf{72.3} & 58.1 & \textbf{52.4} & \textbf{79.7} & 71.1 & 58.9 & \textbf{83.2} & \textbf{67.4} \\
    \bottomrule[1pt]     
    \end{tabular}}
\end{table*}

\begin{figure*}[htbp]
\centering
\subfigure[Adaptability]{
\includegraphics[width=.6\columnwidth]{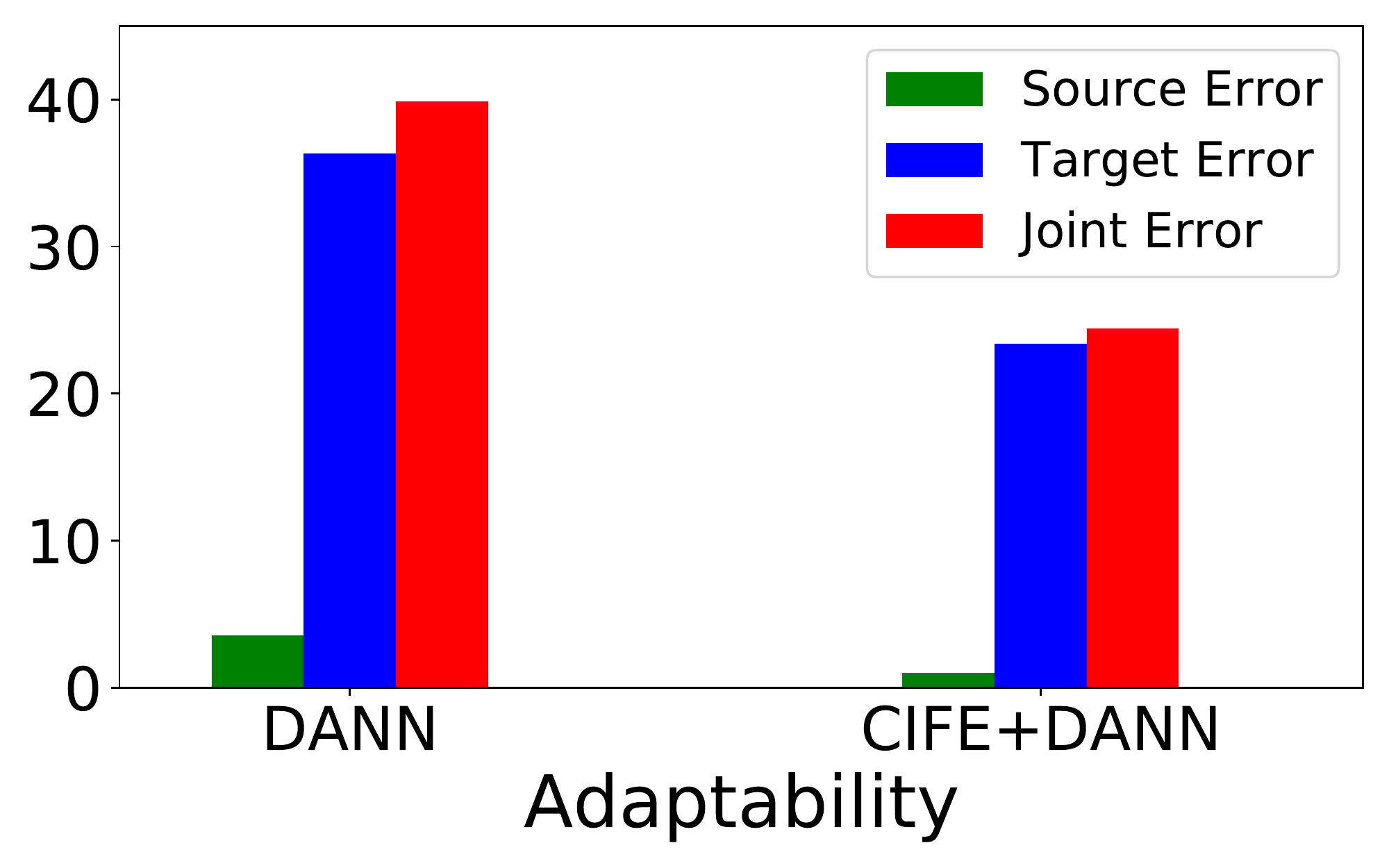}
}
\quad
\subfigure[A-distance]{
\includegraphics[width=.6\columnwidth]{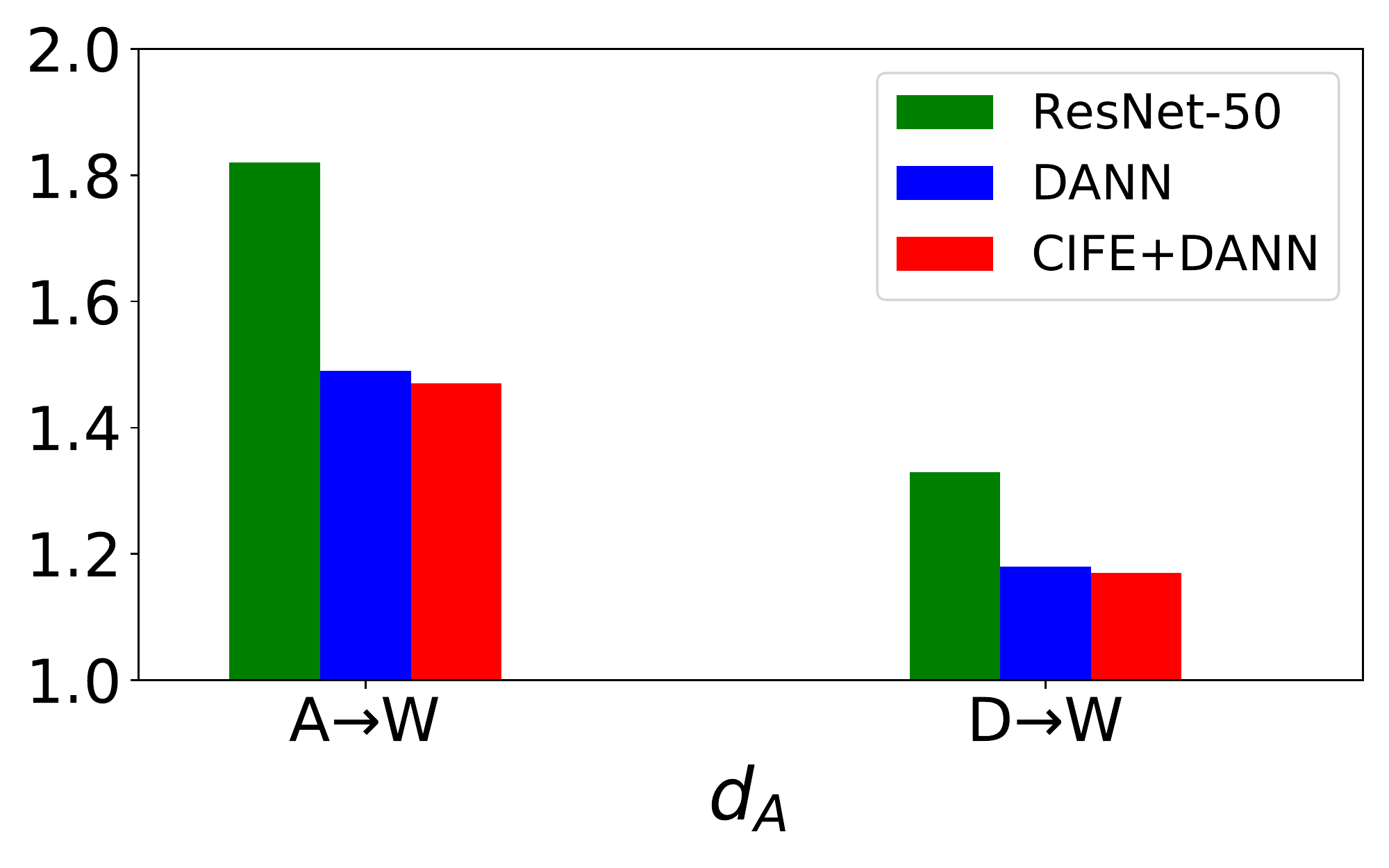}
}
\quad
\subfigure[$\lambda_c$]{
\includegraphics[width=.6\columnwidth]{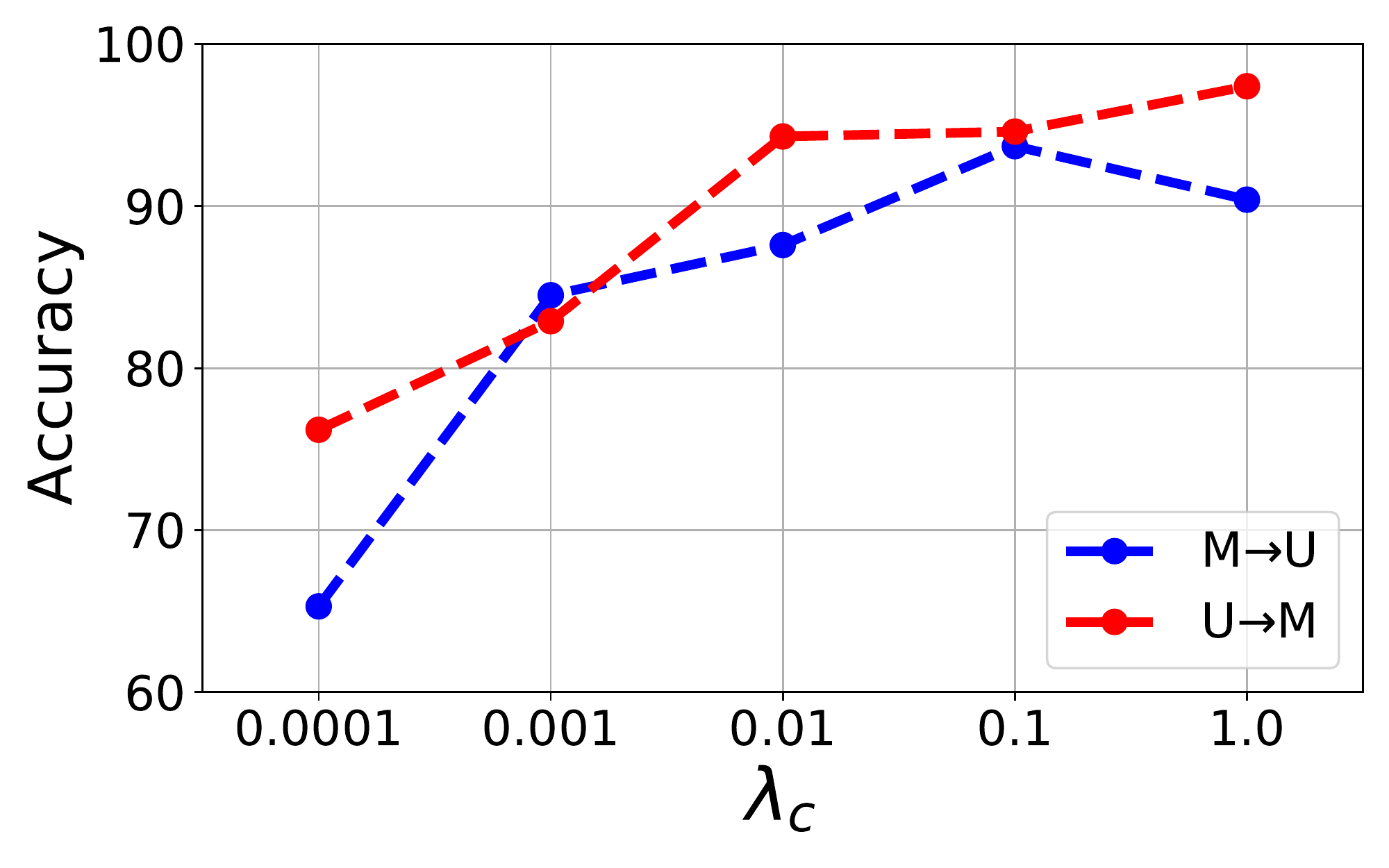}
}
\caption{Analysis of adaptability, domain divergence and parameter sensitivity.}
\label{Fig3}
\end{figure*}

\subsection{Dataset}

\textbf{Office-31} \cite{saenko2010adapting} is a standard domain adaptation dataset. It contains images among 31 classes from 3 domains: Amazon (A) with 2817 images, which contains images downloaded from amazon.com, Webcam (W) with 795 images, and DSLR (D) with 498 images, containing images obtained by web camera and DSLR camera with different settings, respectively. We conduct evaluations on all 6 tasks: A$\to$W, D$\to$W, W$\to$D, A$\to$D, D$\to$A, and W$\to$A

\textbf{ImageCLEF-DA} is a benchmark for ImageCLEF 2014 domain adaptation challenges. It is organized by selecting 12 common classes shared by three domains: Caltech-256 (C), ImageNet ILSVRC 2012 (I), and Pascal VOC 2012 (P). Each domain contains 600 images and 50 images for each class. The three domains in this dataset are of the same size, which is good complementation of the Office-31 dataset where different domains are of different sizes. We evaluate all methods on 6 tasks: I$\to$P, P$\to$I, I$\to$C, C$\to$I, C$\to$P, and P$\to$C.

\textbf{VisDA-2017} \cite{peng2017visda} is a large simulation-to-real dataset with two domains. It contains over 280,000 images of 12 classes. The source domain is termed Synthetic which contains images obtained by rendering 3D models of the same object classes as in the real data from different angles and under different lighting conditions. The target domain is termed Real which comprises natural images. We evaluate the task: Synthetic$\to$Real. 

\textbf{Digits}. We investigate three digits datasets: MNIST, USPS, and Street View House Numbers (SVHN). Each dataset contains digit images of 10 classes (0-9). MNIST consists of grayscale handwritten digit images of size $28\times28$, USPS contains $16\times16$ grayscale images and SVHN composes $32\times32$ colored images which might contain more than one digit in each image. We adopt the experimental settings of CyCADA \cite{hoffman2018cycada} with three tasks: MNIST to USPS (M$\to$U), USPS to MNIST (U$\to$M), and SVHN to MNIST (S$\to$M). All input images should be resized to the size of $32\times32$.

\textbf{Office-Home} \cite{venkateswara2017deep} is a more complicated dataset than Office-31, which consists of around 15500 images from 65 classes in office and home settings. There exist 4 domains in this dataset: Artistic Images (Ar) denotes artistic depictions for object images, Clip Art (Cl) shows picture collection of clipart, Product Images (Pr) presents object images with a clear background and is similar to Amazon category in Office-31, and Real-World Images (Rw) represents object images collected with a regular camera. We establish 12 transfer tasks by using all domain combinations.

\subsection{Comparison Methods}

We extend domain adversarial neural network (DANN) \cite{ganin2016domain}, conditional adversarial domain adaptation (CDAN) \cite{long2018conditional} with the proposed category-invariant feature enhancement (CIFE). We compare with a number of state-of-the-art methods: Deep adaptation network (DAN) \cite{long2015learning}, domain adversarial neural network (DANN) \cite{ganin2016domain}, joint adaptation network (JAN) \cite{long2017deep}, multi-adversarial domain adaptation (MADA) \cite{pei2018multi}, conditional adversarial domain adaptation (CDAN) \cite{long2018conditional}, adversarial discriminative domain adaptation (ADDA) \cite{tzeng2017adversarial}, cycle-consistent adversarial domain adaptation (CyCADA) \cite{hoffman2018cycada}, batch spectral penalization (BSP) \cite{chen2019transferability}, dynamic adversarial domain adaptation (DAAN) \cite{yu2019transfer}, batch nuclear-norm maximization (BNM) \cite{cui2020towards}, enhanced transport distance (ETD) \cite{li2020enhanced} and label propagation with augmented anchors (A$^2$LP) \cite{zhang2020label}.

\subsection{Implementation Details}

The standard evaluation protocols \cite{long2018conditional} of unsupervised domain adaptation are followed in our experiments. All labeled source samples and unlabeled target samples are used in the training stage. In the testing stage, we randomly draw some samples from the training source dataset to provide the category-invariant features. The average classification accuracy based on three random experiments is reported. No data augmentation is used in any of the experiments to allow a fair comparison. For Office-31, ImageCLEF-DA, and Office-Home datasets, we use ResNet-50 \cite{he2016deep} pre-trained on ImageNet \cite{krizhevsky2012imagenet} as the backbone. For VisDA-2017 dataset, we use ResNet-101 \cite{he2016deep} pre-trained on ImageNet \cite{krizhevsky2012imagenet} as the backbone. For digits datasets, we adopt a modified version of Lenet architecture as the base network and train models from scratch. For each backbone network, we use all its layers up to the second last one as the domain-invariant feature extractor $F_s$. The domain-specific feature extractor $F_d$ adopts the same architecture as $F_s$. The classifier $C$ uses a single fully-connected layer whose input dimension should be the sum of the output dimensions of $F_s$ and $F_d$. For domain discriminator $D_d$, we use the same architecture as DANN \cite{ganin2015unsupervised}. The architecture of the category discriminator $D_t$ is similar to that of $D_d$, the only difference lies in the top layer, a softmax layer is used for $D_t$ while a sigmoid layer is used for $D_d$.

We implement all experiments using PyTorch. We adopt mini-batch SGD with momentum of 0.9 and the learning rate annealing strategy as \cite{ganin2016domain}: the learning rate is adjusted by $\eta_p=\frac{\eta_0}{{(1+\theta p)}^\beta}$, where $p$ denotes the process of training epochs that is normalized to be in $[0,1]$, and we set $\eta_0=0.01$, $\theta=10$, $\beta=0.75$, which are optimized to promote convergence and low errors on the source domain. $\lambda_d$ is progressively changed from 0 to 1 by multiplying to $\frac{1-exp(-\delta p)}{1+exp(-\delta p)}$, where $\delta=10$. For all experiments, we select $\lambda_c$ in the range $\{0.0001, 0.001, 0.01, 0.1, 1.0\}$ via tuning on the unlabeled target data.

\subsection{Results}

The results on Office-31 are reported in Table \ref{table1}, with experimental results of baselines directly reported from their original papers wherever available. It indicates that our proposed CIFE mechanism significantly improves the accuracies of DANN \cite{ganin2016domain} and CDAN \cite{long2018conditional}, and achieves state-of-the-art results. Specifically, CIFE+CDAN achieves the best accuracies not only on tasks: A$\to$W, W$\to$D, A$\to$D, and D$\to$A, but also on the average accuracy. Compared with CDAN, there is an obvious improvement in classification accuracies on relatively difficult tasks D$\to$A and W$\to$A where the source domain is quite small. Moreover, as reported in Table \ref{table2}, \ref{table3}, and \ref{table4}, our method can also boost the performance of DANN and CDAN on ImageCLEF-DA, VisDA-2017, Digits and Office-Home. In particular, CIFE+CDAN exceeds baselines on 4 out of 6 domains and achieves the best performance in terms of the average accuracy for ImageCLEF-DA dataset. For Digits datasets, CIFE+CDAN can yield better results on all three tasks, it can also achieve the best average performance compared with other methods. For VisDA-2017 dataset, CIFE+CDAN produces the best average classification accuracy among all comparison methods and outperforms the baseline of ResNet-101 model pre-trained on ImageNet with a great margin. For Office-Home dataset, CIFE+CDAN outperforms other baselines on 9 out of 12 domains and yields the best average classification accuracy. 

\subsection{Analysis}

\textbf{Adaptability.} In this study, we investigate how the category-invariant feature influences the adaptability. In order to compute the adaptability, we train a multi-layer perceptron (MLP) classifier over the feature representations learned by DANN \cite{ganin2015unsupervised} and CIFE+DANN on VisDA-2017. The MLP classifier is trained on all labeled data from both the source and target domains. It should be noted that the target labels are only used in this analysis. When training the MLP classifier, all feature extractors in DANN and CIFE+DANN should be fixed. As shown in Figure \ref{Fig3}(a), we compare the error rates of the ideal joint hypothesis on the source domain, the target domain, and their sum. We observe that the adaptability of the CIFE+DANN is much lower than that of DANN. Obviously, a higher error rate indicates weaker discriminability, leading to poor adaptability as suggested by the domain adaptation theory \cite{ben2010theory}.

\textbf{Distribution Discrepancy.} As shown in the domain adaptation theory \cite{ben2010theory}, the domain discrepancy and adaptability are two important factors that bound the generalization error on the target domain. The A-distance \cite{ben2010theory} is a measure of domain discrepancy, defined as $d_A=2(1-2\epsilon)$, where $\epsilon$ is the error rate of the domain discriminator trained to distinguish source features from target features. In this study, we compare the A-distance of ResNet-50 \cite{he2016deep}, DANN \cite{ganin2015unsupervised}, and CIFE+DANN on two tasks of Office-31 dataset: A$\to$W and D$\to$W. The results are shown in Figure \ref{Fig3}(b). It can be noted that the DANN and CIFE+DANN yield smaller A-distances, indicating the adversarial training can effectively reduce the domain divergence. The A-distance of CIFE+DANN is close to that of DANN on both tasks, revealing that our CIFE improves system performance by optimizing the adaptability rather than further reducing the domain divergence.

\textbf{Parameter Sensitivity Analysis.} In this section, we discuss the sensitivity of CIFE+DANN to the values of the hyperparameter $\lambda_c$. We evaluate the influence of $\lambda_c$ on Digits dataset, especially, the M$\to$U and U$\to$M tasks. $\lambda_c$ is explored in the range $\{0.0001, 0.001, 0.01, 0.1, 1.0\}$. The results are shown in Figure \ref{Fig3}(c). From Figure \ref{Fig3}(c), we observe that the selection of $\lambda_c$ has an influence on the system performance. For the task M$\to$U, with an increase of $\lambda_c$, the accuracy increases rapidly and obtains the best value at $\lambda_c=0.1$. After this point, the further increase of $\lambda_c$ deteriorates the performance. For the task U$\to$M, the accuracy increases from 0.0001 to 1.0 and reaches its best at $\lambda_c=1.0$. This analysis suggests that a properly selected $\lambda_c$ can effectively improve the system performance.

\section{Conclusion}

In this paper, we propose a novel category-invariant feature enhancement (CIFE) mechanism for adversarial domain adaptation. The CIFE incorporates category-invariant features to existing adversarial domain adaptation methods to boost the discriminability of the domain-invariant features. It improves the performance of UDA models by optimizing the adaptability. This approach provides an alternative to the mainstream UDA methods which focus on minimizing the divergence between two domains. We demonstrate that the category-invariant features learned from the source domain can be beneficial to the classification of the target domain. The CIFE approach is general and can be embedded into the existing adversarial domain adaptation methods to boost system performance.

{\small
\bibliographystyle{ieee_fullname}
\bibliography{egbib}

\begin{thebibliography}{10}\itemsep=-1pt

\bibitem{ben2010theory}
Shai Ben-David, John Blitzer, Koby Crammer, Alex Kulesza, Fernando Pereira, and
  Jennifer~Wortman Vaughan.
\newblock A theory of learning from different domains.
\newblock {\em Machine learning}, 79(1-2):151--175, 2010.

\bibitem{ben2007analysis}
Shai Ben-David, John Blitzer, Koby Crammer, and Fernando Pereira.
\newblock Analysis of representations for domain adaptation.
\newblock In {\em Advances in neural information processing systems}, pages
  137--144, 2007.

\bibitem{chen2018multinomial}
Xilun Chen and Claire Cardie.
\newblock Multinomial adversarial networks for multi-domain text
  classification.
\newblock In {\em Proceedings of the 2018 Conference of the North American
  Chapter of the Association for Computational Linguistics: Human Language
  Technologies, Volume 1 (Long Papers)}, pages 1226--1240, 2018.

\bibitem{chen2019transferability}
Xinyang Chen, Sinan Wang, Mingsheng Long, and Jianmin Wang.
\newblock Transferability vs. discriminability: Batch spectral penalization for
  adversarial domain adaptation.
\newblock In {\em International Conference on Machine Learning}, pages
  1081--1090, 2019.

\bibitem{cui2020towards}
Shuhao Cui, Shuhui Wang, Junbao Zhuo, Liang Li, Qingming Huang, and Qi Tian.
\newblock Towards discriminability and diversity: Batch nuclear-norm
  maximization under label insufficient situations.
\newblock In {\em Proceedings of the IEEE/CVF Conference on Computer Vision and
  Pattern Recognition}, pages 3941--3950, 2020.

\bibitem{ganin2015unsupervised}
Yaroslav Ganin and Victor Lempitsky.
\newblock Unsupervised domain adaptation by backpropagation.
\newblock In {\em International conference on machine learning}, pages
  1180--1189. PMLR, 2015.

\bibitem{ganin2016domain}
Yaroslav Ganin, Evgeniya Ustinova, Hana Ajakan, Pascal Germain, Hugo
  Larochelle, Fran{\c{c}}ois Laviolette, Mario Marchand, and Victor Lempitsky.
\newblock Domain-adversarial training of neural networks.
\newblock {\em The Journal of Machine Learning Research}, 17(1):2096--2030,
  2016.

\bibitem{gong2013connecting}
Boqing Gong, Kristen Grauman, and Fei Sha.
\newblock Connecting the dots with landmarks: Discriminatively learning
  domain-invariant features for unsupervised domain adaptation.
\newblock In {\em International Conference on Machine Learning}, pages
  222--230, 2013.

\bibitem{goodfellow2016deep}
Ian Goodfellow, Yoshua Bengio, Aaron Courville, and Yoshua Bengio.
\newblock {\em Deep learning}, volume~1.
\newblock MIT press Cambridge, 2016.

\bibitem{goodfellow2014generative}
Ian Goodfellow, Jean Pouget-Abadie, Mehdi Mirza, Bing Xu, David Warde-Farley,
  Sherjil Ozair, Aaron Courville, and Yoshua Bengio.
\newblock Generative adversarial nets.
\newblock In {\em Advances in neural information processing systems}, pages
  2672--2680, 2014.

\bibitem{he2016deep}
Kaiming He, Xiangyu Zhang, Shaoqing Ren, and Jian Sun.
\newblock Deep residual learning for image recognition.
\newblock In {\em Proceedings of the IEEE conference on computer vision and
  pattern recognition}, pages 770--778, 2016.

\bibitem{hoffman2018cycada}
Judy Hoffman, Eric Tzeng, Taesung Park, Jun-Yan Zhu, Phillip Isola, Kate
  Saenko, Alexei Efros, and Trevor Darrell.
\newblock Cycada: Cycle-consistent adversarial domain adaptation.
\newblock In {\em International conference on machine learning}, pages
  1989--1998. PMLR, 2018.

\bibitem{krizhevsky2012imagenet}
Alex Krizhevsky, Ilya Sutskever, and Geoffrey~E Hinton.
\newblock Imagenet classification with deep convolutional neural networks.
\newblock In {\em Advances in neural information processing systems}, pages
  1097--1105, 2012.

\bibitem{li2020enhanced}
Mengxue Li, Yi-Ming Zhai, You-Wei Luo, Peng-Fei Ge, and Chuan-Xian Ren.
\newblock Enhanced transport distance for unsupervised domain adaptation.
\newblock In {\em Proceedings of the IEEE/CVF Conference on Computer Vision and
  Pattern Recognition}, pages 13936--13944, 2020.

\bibitem{long2015learning}
Mingsheng Long, Yue Cao, Jianmin Wang, and Michael Jordan.
\newblock Learning transferable features with deep adaptation networks.
\newblock In {\em International conference on machine learning}, pages 97--105.
  PMLR, 2015.

\bibitem{long2018conditional}
Mingsheng Long, Zhangjie Cao, Jianmin Wang, and Michael~I Jordan.
\newblock Conditional adversarial domain adaptation.
\newblock In {\em Advances in Neural Information Processing Systems}, pages
  1640--1650, 2018.

\bibitem{long2017deep}
Mingsheng Long, Han Zhu, Jianmin Wang, and Michael~I Jordan.
\newblock Deep transfer learning with joint adaptation networks.
\newblock In {\em International conference on machine learning}, pages
  2208--2217. PMLR, 2017.

\bibitem{pan2009survey}
Sinno~Jialin Pan and Qiang Yang.
\newblock A survey on transfer learning.
\newblock {\em IEEE Transactions on knowledge and data engineering},
  22(10):1345--1359, 2009.

\bibitem{pei2018multi}
Zhongyi Pei, Zhangjie Cao, Mingsheng Long, and Jianmin Wang.
\newblock Multi-adversarial domain adaptation.
\newblock {\em arXiv preprint arXiv:1809.02176}, 2018.

\bibitem{peng2017visda}
Xingchao Peng, Ben Usman, Neela Kaushik, Judy Hoffman, Dequan Wang, and Kate
  Saenko.
\newblock Visda: The visual domain adaptation challenge.
\newblock {\em arXiv preprint arXiv:1710.06924}, 2017.

\bibitem{saenko2010adapting}
Kate Saenko, Brian Kulis, Mario Fritz, and Trevor Darrell.
\newblock Adapting visual category models to new domains.
\newblock In {\em European conference on computer vision}, pages 213--226.
  Springer, 2010.

\bibitem{sun2016deep}
Baochen Sun and Kate Saenko.
\newblock Deep coral: Correlation alignment for deep domain adaptation.
\newblock In {\em European conference on computer vision}, pages 443--450.
  Springer, 2016.

\bibitem{tzeng2017adversarial}
Eric Tzeng, Judy Hoffman, Kate Saenko, and Trevor Darrell.
\newblock Adversarial discriminative domain adaptation.
\newblock In {\em Proceedings of the IEEE conference on computer vision and
  pattern recognition}, pages 7167--7176, 2017.

\bibitem{venkateswara2017deep}
Hemanth Venkateswara, Jose Eusebio, Shayok Chakraborty, and Sethuraman
  Panchanathan.
\newblock Deep hashing network for unsupervised domain adaptation.
\newblock In {\em Proceedings of the IEEE Conference on Computer Vision and
  Pattern Recognition}, pages 5018--5027, 2017.

\bibitem{wu2020dual}
Yuan Wu and Yuhong Guo.
\newblock Dual adversarial co-learning for multi-domain text classification.
\newblock In {\em Proceedings of the AAAI Conference on Artificial
  Intelligence}, volume~34, pages 6438--6445, 2020.

\bibitem{wu2020dualb}
Yuan Wu, Diana Inkpen, and Ahmed El-Roby.
\newblock Dual mixup regularized learning for adversarial domain adaptation.
\newblock In {\em European Conference on Computer Vision}, pages 540--555.
  Springer, 2020.

\bibitem{wu2021conditional}
Yuan Wu, Diana Inkpen, and Ahmed El-Roby.
\newblock Conditional adversarial networks for multi-domain text
  classification.
\newblock In {\em Proceedings of the Second Workshop on Domain Adaptation for
  NLP}, pages 16--27, 2021.

\bibitem{wu2021mixup}
Yuan Wu, Diana Inkpen, and Ahmed El-Roby.
\newblock Mixup regularized adversarial networks for multi-domain text
  classification.
\newblock In {\em ICASSP 2021-2021 IEEE International Conference on Acoustics,
  Speech and Signal Processing (ICASSP)}, pages 7733--7737. IEEE, 2021.

\bibitem{yan2017mind}
Hongliang Yan, Yukang Ding, Peihua Li, Qilong Wang, Yong Xu, and Wangmeng Zuo.
\newblock Mind the class weight bias: Weighted maximum mean discrepancy for
  unsupervised domain adaptation.
\newblock In {\em Proceedings of the IEEE Conference on Computer Vision and
  Pattern Recognition}, pages 2272--2281, 2017.

\bibitem{yu2019transfer}
Chaohui Yu, Jindong Wang, Yiqiang Chen, and Meiyu Huang.
\newblock Transfer learning with dynamic adversarial adaptation network.
\newblock In {\em 2019 IEEE International Conference on Data Mining (ICDM)},
  pages 778--786. IEEE, 2019.

\bibitem{zhang2020label}
Yabin Zhang, Bin Deng, Kui Jia, and Lei Zhang.
\newblock Label propagation with augmented anchors: A simple semi-supervised
  learning baseline for unsupervised domain adaptation.
\newblock In {\em European Conference on Computer Vision}, pages 781--797.
  Springer, 2020.

\end{thebibliography}
}

\end{document}